\documentclass{article}
\usepackage{spconf,amsmath,epsfig}
\usepackage{cite}
\usepackage{float}
\usepackage{amsmath,amssymb,amsfonts}
\usepackage{graphicx}
\usepackage{textcomp}
\usepackage{xcolor}
\usepackage{algorithmic,algorithm}
\usepackage{booktabs}

\title{Towards Adaptive Semantic Segmentation by Progressive Feature Refinement}
\name{Bin Zhang$^{\star \dagger}$ \qquad Shengjie Zhao$^{\star \dagger}$ \qquad Rongqing Zhang$^{\star}$}
\address{$^{\star}$School of Software Engineering, Tongji University, Shanghai, China\\
	$^{\dagger}$Key Laboratory of Embedded System and Service Computing, Tongji University, Shanghai, China\\
   }
\begin{document}
%
\maketitle
\begin{abstract}
As one of the fundamental tasks in computer vision, semantic segmentation plays an important role in real world applications. Although numerous deep learning models have made notable progress on several mainstream datasets with the rapid development of convolutional networks, they still encounter various challenges in practical scenarios. Unsupervised adaptive semantic segmentation aims to obtain a robust classifier trained with source domain data, which is able to maintain stable performance when deployed to a target domain with different data distribution. In this paper, we propose an innovative progressive feature refinement framework, along with domain adversarial learning to boost the transferability of segmentation networks. Specifically, we firstly align the multi-stage intermediate feature maps of source and target domain images, and then a domain classifier is adopted to discriminate the segmentation output. As a result, the segmentation models trained with source domain images can be transferred to a target domain without significant performance degradation. Experimental results verify the efficiency of our proposed method compared with state-of-the-art methods.
\end{abstract}
\section{Introduction}

Recent advances in deep learning \cite{He_2016_CVPR} have revolutionized the development of many computer vision tasks, such as person re-identification, depth estimation, semantic segmentation, etc. Semantic segmentation can be regarded as a dense prediction task, which aims to assign a category label (e.g. building, sidewalk, bus, train) to each image pixel. Due to the emergence of diverse large scale datasets  \cite{Cordts_2016_CVPR},  a wide variety of deep learning models \cite{Chen2018DeepLab, Sun_2019_CVPR, Chen_2018_ECCV, Zhao_2017_CVPR} have gained remarkable breakthrough. Nevertheless, collecting high-resolution and well-annotated datasets is a laborious process, especially for tasks that require pixel-level annotation. An alternative solution is to take the advantage of synthetic data collected from simulated environment where unlimited amount of computer-generated images are available.
\begin{figure}[htbp]
	\centerline{\includegraphics[width=245pt,height=180pt]{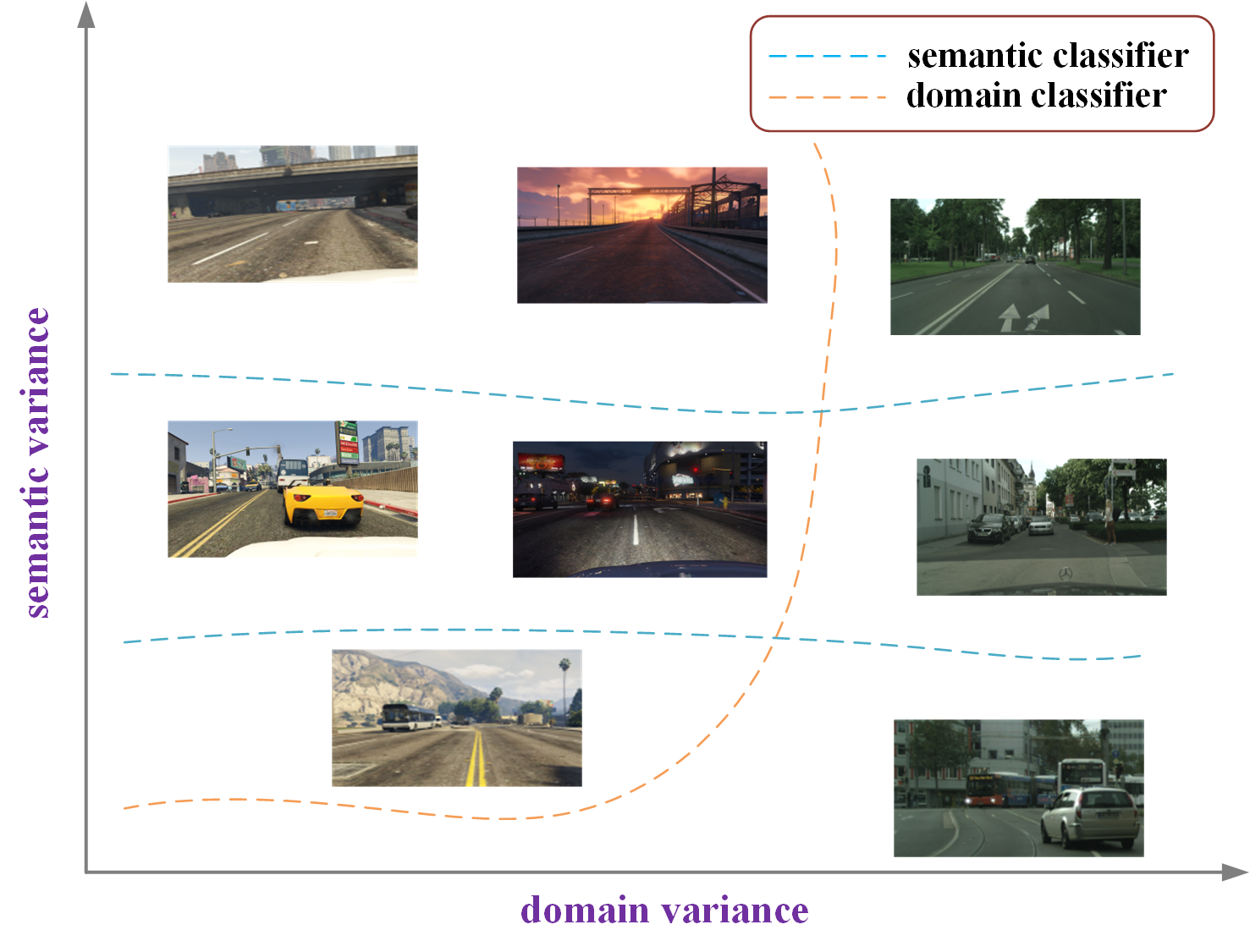}}
	\caption{Complementary utilization of semantic classifier and domain classifier for adaptive segmentation.}
	\label{fig1}
\end{figure} 
Despite this, generic segmentation models are commonly domain-specific and thus inevitably suffer from the dataset bias problem \cite{Torralba2011Unbiased}, which hinders their cross-domain generalization to novel scenes. Therefore, how to improve the adaptation ability of deep learning models is of great significance.

Unsupervised domain adaptation offers a formal framework for addressing the above-mentioned issues by bridging the domain gap between the source and target domains. The core idea is transferring domain-invariant knowledge from a source domain with sufficient labels to a target domain without annotation. Among a majority of previous approaches \cite{Zhang_2018_CVPR, Hoffman2018CyCADA}, they either minimize the difference between intermediate feature distribution of source and target data via adversarial learning, or explicitly transfer source domain data into target domain in the input space. For low-level vision tasks such as image classification, the feature maps extracted by deep convolutional neural networks are aligned across source and target domains. However, these methods often fail when handling high-level vision tasks such as semantic segmentation which encodes complicated relationship among diverse object categories.

To this end, we propose an innovative learning methodology for cross-domain semantic segmentation, termed progressive feature refinement, which disentangles the style and content representation of source and target domain images, respectively. In this way, the source and target features can be aligned stage by stage and the obtained feature maps are less sensitive to domain shift. Our basic concept is illustrated in Figure 1. In addition to the conventional segmentation network, we also borrow the experience from \cite{Tsai2018Learning} and exploit a domain classifier to obtain domain-invariant output. The proposed framework can be trained end-to-end in an adversarial learning manner. Evaluations on mainstream benchmarks demonstrate that our approach is superior to most state-of-the-art baselines. 

\section{Related Work}
\subsection{Domain Adaptation}
Semantic segmentation has always been one of the research hotspots in computer vision and is valuable for a large number of applications, including autonomous driving, robot scene understanding, medical image analysis, etc. Powered by high-capacity deep neural networks, \cite{Chen2018DeepLab} equipped the ResNet-101 with spatial pyramid pooling module and reaches high segmentation accuracy. In order to enlarge the receptive field and retain high resolution of feature maps, \cite{Yu2016Multi} aggregated multi-scale context information through dilated convolution. Although traditional deep learning models have been proven to be effective on the segmentation task, their configuration design is non-trivial. Most recently, \cite{Nekrasov_2019_CVPR} adopted neural architecture search (NAS) strategy and reinforcement learning to discover the optimal network structure automatically rather than rely on the tedious manual design.

\subsection{Domain Adaptation for Semantic Segmentation}
While semantic segmentation is a well-researched topic, few efforts have been made to explore the adaptation of segmentation models. In \cite{Hoffman2018CyCADA}, they employed image-to-image translation networks to convert source domain images into target style, followed by adversarial learning to ensure the extracted feature maps are domain-agnostic. By introducing the concept of domain flow, \cite{Gong_2019_CVPR} generated a series of intermediate domains and smoothly mitigated the domain gap. In \cite{choi2019selfensembling}, they combined GAN-based augmentation strategy with self-ensembling techniques and produced augmented labeled images by mimicking the target domain style. Different from the above global alignment method which ignores semantic consistency, \cite{Luo_2019_CVPR} took the category-level information into consideration and alleviated the negative transfer problem. In \cite{Ramirez_2019_ICCV}, they proved that most visual tasks are closely related to each other and proposed a unified learning scheme to investigate the latent relationship across distinct tasks and domains. Apart from domain adaptation, \cite{Yue_2019_ICCV} also tackled a more difficult domain generalization case, where both data and label of target domain are unavailable.

\section{Proposed Method}
In this paper, we focus on the unsupervised cross-domain semantic segmentation setting, where we are given a labeled source domain dataset $\mathcal{D}_{s}$ with pixel-level annotation $\mathcal{Y}_{s}$ and an unlabeled target domain dataset $\mathcal{D}_{t}$. Our goal is to utilize the source dataset $\mathcal{D}_{s}$ to train a model $M$ that can precisely provide segmentation map for images in $\mathcal{D}_{t}$. Figure 2 depicts the overall architecture of our proposed framework, which is composed of the semantic segmentation network, the progressive feature refinement module (PFR), and the domain adversarial learning module in the output space.

\subsection{Semantic Segmentation Network}
We adopt the same setting as the method in \cite{Tsai2018Learning} and utilize the DeepLab-v2 network with pretrained ResNet-101 \cite{He_2016_CVPR} backbone as our base model. We discard the last fully connected layer and modify the strides of the last two convolution layers to 1. Given an image $x_{s} \in \mathbb{R}^{H \times W \times 3}$ with height $H$ and width $W$ from the source domain dataset $\mathcal{D}_{s}$, the semantic segmentation network $M$ is optimized to produce the segmentation output $p_{s}=M\left(x_{s}\right) \in \mathbb{R}^{H \times W \times C}$ for $C$ different categories. This is accomplished by minimizing the following segmentation loss $\mathcal{L}_{seg}$ under the supervision of the corresponding ground truth label map $y_{s} \in \mathcal{Y}_{s}$:
\begin{equation}
\mathcal{L}_{seg}=-\sum_{h=1}^{H} \sum_{w=1}^{W} \sum_{c=1}^{C} y_{s}^{(h, w, c)} \log p_{s}^{(h, w, c)}
\end{equation}

\subsection{Progressive Feature Refinement}
To align the feature maps of source and target domain data, we insert the additional PFR module to the existing semantic segmentation network. We are inspired by the fact that the style and content of an image are separable \cite{Huang_2017_ICCV}, and our main idea is to disentangle the style and content feature representation of source and target domain images in order to align them progressively. 

More formally, let $C_{i}^{s}$ and $C_{i}^{t}$ denote the content feature of source and target domain image obtained from the $i$-th stage ($i=2,3,4,5$) of the backbone ResNet-101, respectively. Similarly, we use $S_{i}^{s}$ and $S_{i}^{t}$ to represent the source and target style feature, i.e., the Gram matrix \cite{Gatys_2016_CVPR} calculated by the content feature. We define the following content loss $\mathcal{L}_{con}$ and style loss $\mathcal{L}_{sty}$ to match the style and content feature of source domain images to those of target domain images:
\begin{equation}
\mathcal{L}_{sty}=\sum_{i}\left\|S_{i}^{s}-S_{i}^{t}\right\|_{2}
\end{equation}
\begin{equation}
\mathcal{L}_{con}=\sum_{i}\left\|C_{i}^{s}-C_{i}^{t}\right\|_{2}
\end{equation} 
The training
\begin{figure*}[!h]
	\centerline{\includegraphics[width=485pt,height=220pt]{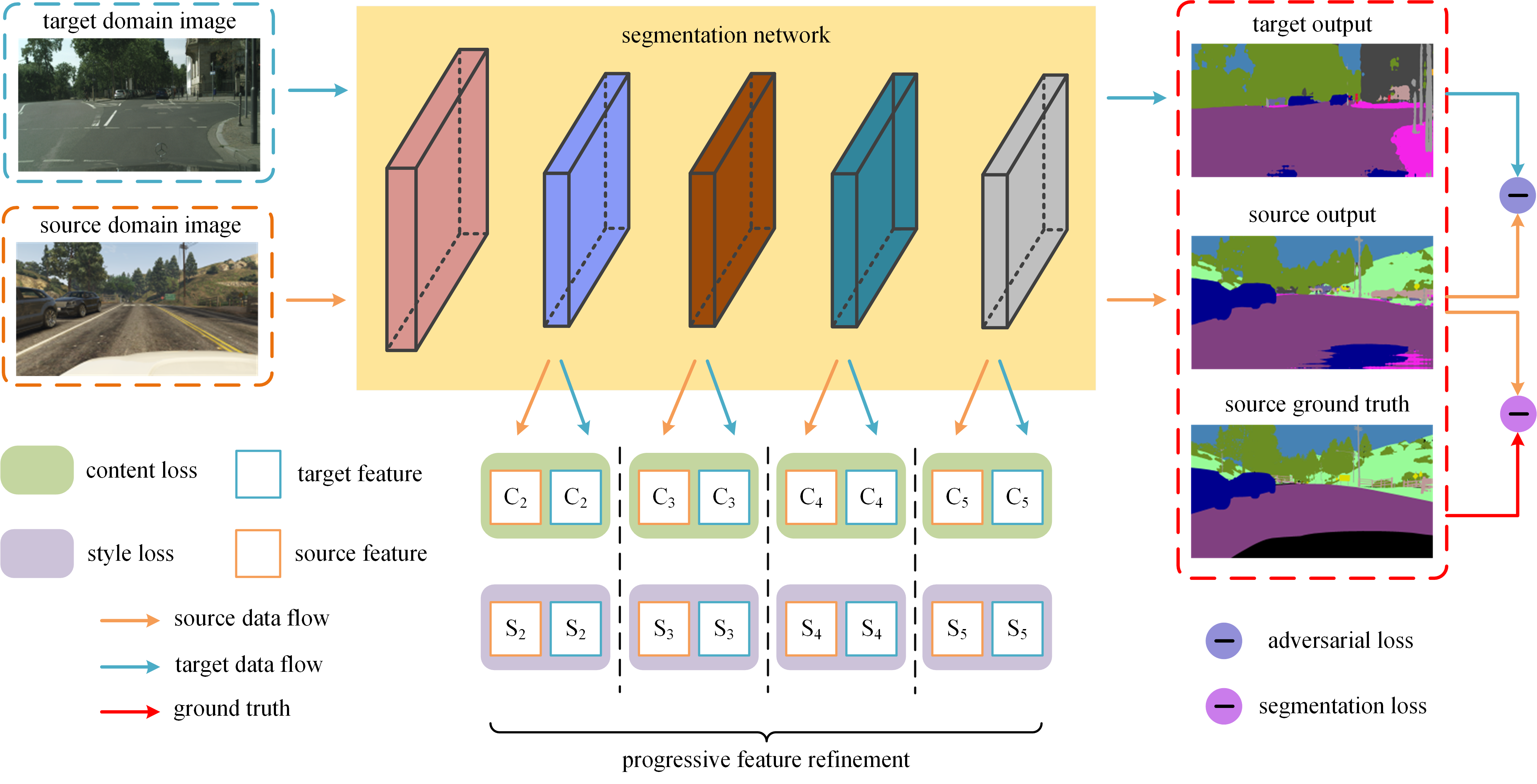}}
	\caption{The overall pipeline of our adaptive semantic segmentation  framework. We augment the general semantic segmentation network with the progressive feature refinement module, and incorporate domain adversarial learning to further improve the segmentation result in the output space.}
	\label{fig2}
\end{figure*}
objective of PFR module can be summarized as follows:
\begin{equation}
\mathcal{L}_{pfr}=\mathcal{L}_{sty}+\mathcal{L}_{con}
\end{equation} 
which is a combination of two parts, i.e., the style loss $\mathcal{L}_{sty}$ and the content loss $\mathcal{L}_{con}$.

\subsection{Domain Adversarial Learning}
On the one hand, the segmentation loss $\mathcal{L}_{seg}$ is optimized with the source domain images and thus has no contribution to narrow the domain discrepancy. On the other hand, the PFR module only eliminates the domain variance in the feature space and there is no guarantee of domain-agnostic output.
Therefore, we further integrate domain adversarial learning to rectify the segmentation results. We apply an auxiliary domain classifier $D$ to the predictions of both source and target domain images and force $D$ to discriminate whether the segmentation output is from source domain or target domain. The adversarial loss is defined as:
\begin{equation}
\begin{aligned} 
\mathcal{L}_{adv} &=\mathbb{E}_{x \sim \mathcal{D}_{s}}\left[\log D\left(M(x)\right)\right] \\ &+\mathbb{E}_{x \sim \mathcal{D}_{t}}\left[\log \left(1-D\left(M(x)\right)\right)\right]
\end{aligned}
\end{equation}

\subsection{Network Optimization}
We combine the aforementioned loss functions and formulate the overall training objective of our framework as:
\begin{equation}
\mathcal{L}_{total}=\mathcal{L}_{seg}+\lambda_{adv} \mathcal{L}_{adv}+\lambda_{pfr} \mathcal{L}_{pfr}
\end{equation}
where $\lambda_{adv}$ and $\lambda_{pfr}$ are adjustable hyper-parameters that control the trade-off among regularization terms. In our implementation we set $\lambda_{adv}=0.002$ and $\lambda_{pfr}=0.004$.

\section{Experiments}
\subsection{Datasets and Evaluation Metric}
We verify the performance of our proposed approach on the GTA5 \cite{Richter2016Playing} $\rightarrow$ Cityscapes \cite{Cordts_2016_CVPR} domain adaptation tasks. Cityscapes is a large-scale dataset to evaluate the accuracy of semantic segmentation models, which covers the urban scenes of several European countries. It is split into a training set with 2,975 samples, a testing set with 1,525 samples, and a validation set with 500 samples. GTA5 dataset contains 24,966 high-definition images collected from a contemporary computer game called Grand Theft Auto V. The dataset is automatically annotated into 19 categories, which are consistent with the Cityscapes dataset. As for the evaluation metric, we choose the commonly adopted Intersection over Union (IoU) for fair comparison:
\begin{equation}
IoU=\frac{TP}{TP+FP+FN}
\end{equation}
where TP, FP, FN stand for the number of true positives, false positives, and false negatives, respectively.
\renewcommand\tabcolsep{0.7pt}
\renewcommand\arraystretch{1.0}
\begin{table*}[!h]
	\caption{Performance comparison between baseline approaches and ours under the GTA5 $\rightarrow$ Cityscapes setting.}
	\begin{tabular}{c|c|ccccccccccccccccccc|c}
		\toprule[1pt]
		\hline
		Methods     & Backbone & \rotatebox{90}{road} & \rotatebox{90}{sidewalk} & \rotatebox{90}{building} & \rotatebox{90}{wall} & \rotatebox{90}{fence} & \rotatebox{90}{pole} & \rotatebox{90}{t light} & \rotatebox{90}{t sign} & \rotatebox{90}{veg}  & \rotatebox{90}{terrain} & \rotatebox{90}{sky}  & \rotatebox{90}{person} & \rotatebox{90}{rider} & \rotatebox{90}{car}  & \rotatebox{90}{truck} & \rotatebox{90}{bus}  & \rotatebox{90}{train} & \rotatebox{90}{motorbike} & \rotatebox{90}{bicycle} & \textbf{mIoU} \\ \hline
		FCN WId \cite{hoffman2016fcns}     & VGG-16      & 70.4 & 32.4     & 62.1     & 14.9 & 5.4   & 10.9 & 14.2    & 2.7    & 79.2 & 21.3    & 64.6 & 44.1   & 4.2   & 70.4 & 8.0   & 7.3  & 0.0   & 3.5       & 0.0  & 27.1 \\ \hline
		Curriculum \cite{Yang2017Curriculum} & VGG-16      & 72.9 & 30.0     & 74.9     & 12.1 & 13.2  & 15.3 & 16.8    & 14.1   & 79.3 & 14.5    & 75.5 & 35.7   & 10.0  & 62.1 & 20.6  & 19.0 & 0.0   & 19.3      & 12.0 & 31.4 \\ \hline
		TGCF-DA \cite{choi2019selfensembling} & VGG-16      & 90.2 & \textbf{51.5}     & 81.1     & 15.0 & 10.7  & 37.5 & 35.2    & 28.9   & \textbf{84.1} & 32.7    & 75.9 & 62.7   & 19.9  & \textbf{82.6} & 22.9  & 28.3 & 0.0   & 19.3      & 12.0 & 42.5 \\ \hline
		ROAD \cite{Chen_2018_CVPR}  & VGG-16      & 85.4 & 31.2     & 78.6     & 27.9 & 22.2  & 21.9 & 23.7    & 11.4   & 80.7 & 29.3    & 68.9 & 48.5   & 14.1  & 78.0 & 19.1  & 23.8 & \textbf{9.4}   & 8.3       & 0.0  & 35.9 \\ \hline
		Cycada \cite{Hoffman2018CyCADA} & ResNet-101   & 86.7 & 35.6     & 80.1     & 19.8 & 17.5  & \textbf{38.0} & \textbf{39.9}    & \textbf{41.5}   & 82.7 & 27.9    & 73.6 & \textbf{64.9}   & 19.0  & 65.0 & 12.0  & 28.6 & 4.5   & 31.1      & \textbf{42.0} & 42.7 \\ \hline
		CLAN \cite{Luo_2019_CVPR} & ResNet-101   & 87.0 & 27.1     & 79.6     & 27.3 & 23.3  & 28.3 & 35.5    & 24.2   & 83.6 & 27.4    & 74.2 & 58.6   & 28.0  & 76.2 & 33.1  & 36.7 & 6.7   & \textbf{31.9}      & 31.4 & 43.2 \\ \hline
		AdaptSegNet \cite{Tsai2018Learning}& ResNet-101   & 86.5 & 36.0     & 79.9     & 23.4 & 23.3  & 23.9 & 35.2    & 14.8   & 83.4 & 33.3    & 75.6 & 58.5   & 27.6  & 73.7 & 32.5  & 35.4 & 3.9   & 30.1      & 28.1 & 42.4 \\ \hline
		DLOW \cite{Gong_2019_CVPR}  & ResNet-101   & 87.1 & 33.5    & 80.5    & 24.5 & 13.2  & 29.8 & 29.5    & 26.6   & 82.6 & 26.7   & \textbf{81.8} & 55.9   & 25.3  & 78.0 & 33.5  & 38.7 & 0.0  & 22.9      & 34.5 & 42.3 \\ \hline
		Ours        & ResNet-101   & \textbf{90.8} & 40.8    & \textbf{81.9}     & \textbf{28.4}  & \textbf{24.4}  & 24.2 & 32.0    & 17.6   & 83.8 & \textbf{36.6}    & 72.4 & 59.3   & \textbf{29.0}  & 82.1 & \textbf{35.5}   & \textbf{45.6}  & 3.3   & 29.1     & 28.6 & \textbf{44.5} \\ \hline
		\bottomrule[1pt]
	\end{tabular}
\end{table*}
\begin{figure*}[!h]
	\centerline{\includegraphics[width=500pt,height=221pt]{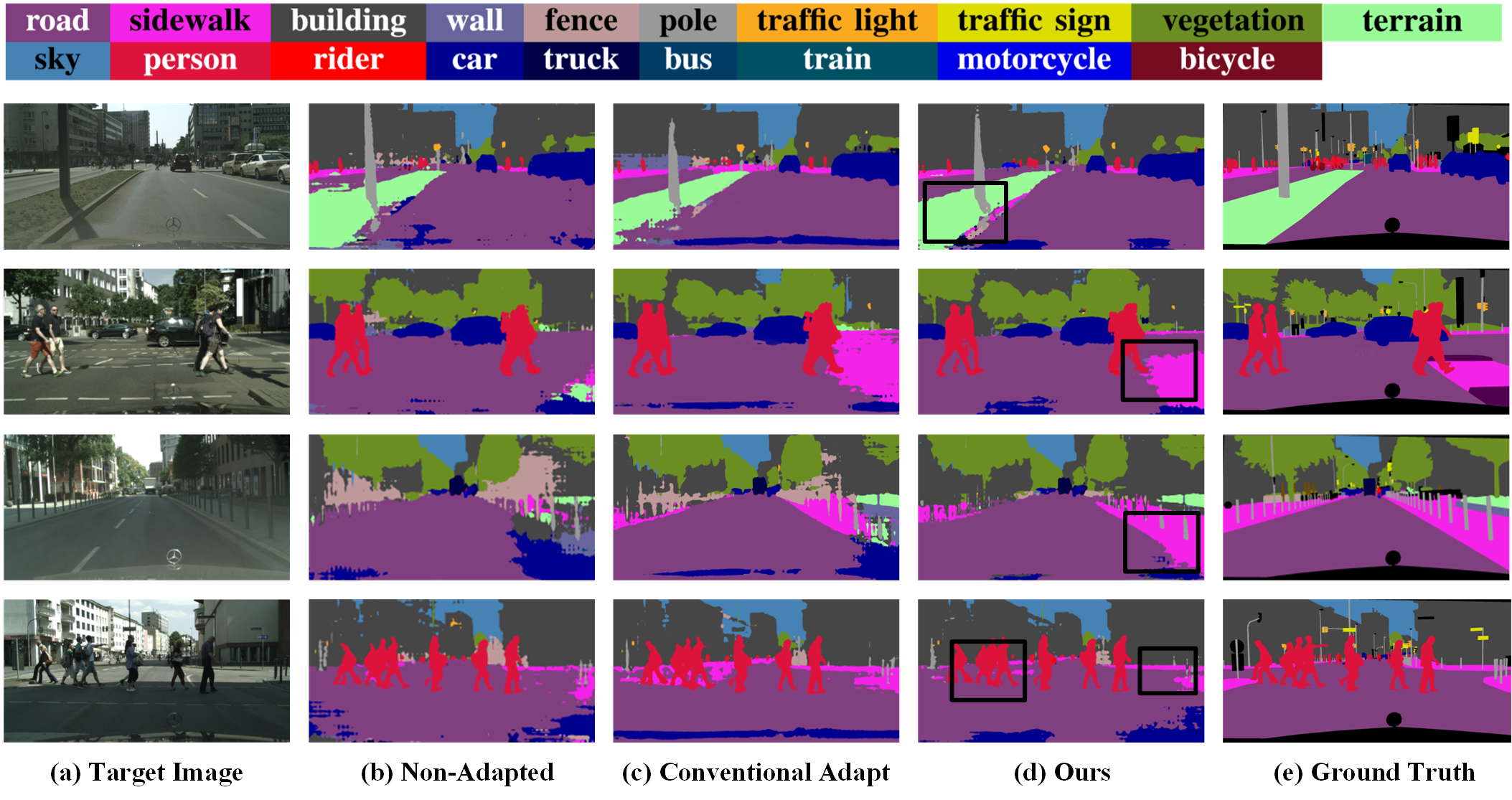}}
	\caption{Qualitative segmentation results on the GTA5 $\rightarrow$ Cityscapes domain adaptation task. We present (a) target image, (b) Non-Adapted, (c) Conventional Adapt \cite{Tsai2018Learning}, (d) Ours, and (e) Ground Truth. Details are highlighted by the black boxes.}
	\label{fig4}
\end{figure*}
\subsection{Performance Comparison}
We take the GTA5 dataset as source domain and the Cityscapes dataset as target domain. Table 1 summarizes the comparison results between baseline approaches and ours on the GTA5 $\rightarrow$ Cityscapes domain adaptation task. As presented in Table 1, equipped with the ResNet-101 backbone, our approach reaches the best mIoU result compared to other baseline methods. By further looking into the results of each category, we observe that the improvement over other methods primarily comes from the ``road'', ``building'', ``wall'', ``fence'', ``terrain'', ``rider'', ``truck'', and ``bus'' classes. 
Over the input space alignment baselines which translate the source domain images into target style \cite{Hoffman2018CyCADA}, our PFR is more memory-efficient since there is no requirement for extra image-to-image translation networks. Although PFR achieves limited performance gain on some less frequent objects which may trigger a negative transfer compared to class-wise alignment methods \cite{Luo_2019_CVPR}, it is excellent at the dominant classes such as ``road'', ``building'', etc. Moreover, we visualize some qualitative segmentation examples and their corresponding ground truth in Figure 3. It is obvious that our method can segment the object boundaries more precisely and produce smoother output.

\section{Conclusions}
In this paper, we propose a novel progressive feature refinement method 
for cross-domain semantic segmentation. Our proposed PFR provides a novel perspective of insight  
by incorporating the content and style alignment module. 
The experimental results demonstrate that PFR outperforms most current state-of-the-art unsupervised domain adaptation methods on the GTA5 $\rightarrow$ Cityscapes task. 
\\
\\
\textbf{Acknowledgement.} This work is supported in part by the National Key Research and Development Project under Grant 2019YFB2102300 and 2019YFB2102301, and in part by the National Natural Science Foundation of China under Grant 61936014 and 61901302.

\bibliographystyle{IEEEbib}
\bibliography{ref}

\end{document}